\RequirePackage{fix-cm}

\documentclass[conference,letterpaper]{IEEEtran}
\IEEEoverridecommandlockouts

\usepackage{cite}
\usepackage{amsmath,amssymb,amsfonts}
\usepackage{graphicx}
\makeatletter
\ifdefined\Gread@transgroupfalse
  \AddToHook{cmd/Gread@@pdftex/before}{\Gread@transgroupfalse}
\fi
\makeatother
\usepackage{pgf}
\usepackage{helvet}
\usepackage{microtype}
\usepackage{textcomp}
\usepackage[table]{xcolor}
\usepackage{booktabs}
\usepackage{array}
\usepackage{float}
\usepackage[colorlinks=true,
            linkcolor=black,
            citecolor=black,
            urlcolor=blue,
            breaklinks=true]{hyperref}
\usepackage{arydshln}
\graphicspath{{img/}}

\def\BibTeX{{\rm B\kern-.05em{\sc i\kern-.025em b}\kern-.08em
    T\kern-.1667em\lower.7ex\hbox{E}\kern-.125emX}}

\begin{document}
\raggedbottom

\bstctlcite{IEEEexample:BSTcontrol}

\title{Predicted Futures Are Not Enough: Learning Executable Goals for Robot Manipulation}
\author{%
\IEEEauthorblockN{Tzu-Yu Chuang\IEEEauthorrefmark{1},
Ching-Hsiang Chang\IEEEauthorrefmark{2},
Yi-Hsiu Lee\IEEEauthorrefmark{2},
Yi-Ting Chen\IEEEauthorrefmark{1},
Min Sun\IEEEauthorrefmark{2},
YuanFu Yang\IEEEauthorrefmark{1}}
\IEEEauthorblockA{\IEEEauthorrefmark{1}National Yang Ming Chiao Tung University\qquad
\IEEEauthorrefmark{2}National Tsing Hua University}
}
\maketitle

\begin{abstract}
Generative world models provide rich predictions of how manipulation scenes may evolve toward task objectives, yet those futures do not directly expose the compact task variables required by control. When training supervises future prediction alone, terminal goal accuracy is not an explicit learning objective, even when geometric recovery is available. We present \emph{Entity-Level Goal Readout}, a learned prediction-to-execution interface that makes the executable terminal goal an explicit output of a 3D trace world model. It combines object-centric pose prediction with translation grounded in observed depth to produce a compact goal in $SE(3)$. A shared Pose-Native Executor consumes this fixed goal with online object-pose feedback for closed-loop control without rerunning the world model. Across five manipulation tasks, the pipeline achieves a mean success rate of $79.69\%$. Goal diagnostics directly measure terminal goal accuracy, while controlled translation perturbations characterize how execution degrades under goal error. Zero-shot deployment on a Franka arm achieves $73.33\%$ success on nominal StackCube, $66.67\%$ with distractors, and $75.00\%$ on PickPlate with a target unseen during policy training. These results support treating the prediction-to-execution interface as an explicit learned component of world-model planning rather than incidental post-processing in the control pipeline itself.
Project page: \url{https://claire0730.github.io/executable-goals/}
\end{abstract}

\section{Introduction}
\label{sec:intro}


\begin{figure}[!t]
\centering
\includegraphics[
    width=\columnwidth,
    trim=324bp 25bp 346bp 32bp,
    clip
]{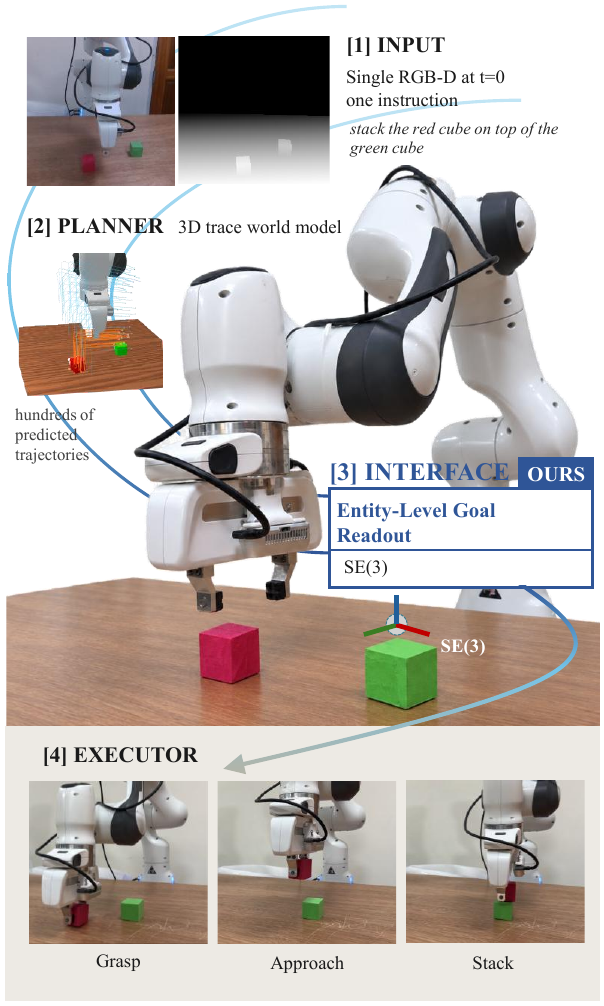}
\caption{\textbf{From prediction to an executable goal.} A single RGB-D observation and task instruction condition a 3D trace world model. The Entity-Level Goal Readout converts the resulting world-model representation into a compact object-relative goal in $SE(3)$ for the Pose-Native Executor. The StackCube example illustrates the prediction-to-execution interface from a rich predicted future to closed-loop manipulation.}
\label{fig:teaser}
\end{figure}


\begin{figure*}[!t]
\centering
\includegraphics[
    width=\textwidth,
    trim=0bp 115bp 0bp 165bp,
    clip
]{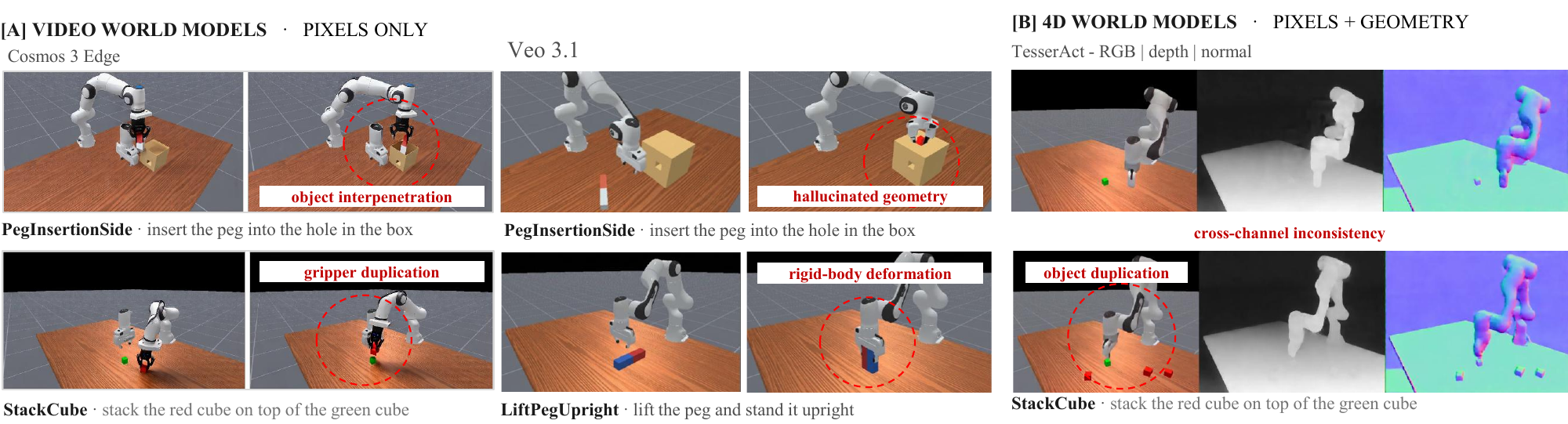}
\caption{\textbf{Qualitative motivation.} Cosmos~3 Edge~\cite{cosmos3_2026}, Veo~3.1~\cite{veo31_2026}, and TesserAct~\cite{tesseract2025} illustrate interpenetration, hallucinated geometry, duplicated objects or grippers, deformation, and inconsistent RGB, depth, and normals. These prediction defects motivate evaluating the control goals recovered from predicted futures. The separate goal diagnostic in Table~\ref{tab:goalerr} excludes Veo~3.1.}
\label{fig:motivation}
\end{figure*}

Generative world models support robot manipulation by predicting how scenes can evolve toward task objectives. Yet rich future predictions do not by themselves expose the compact task variables required by control. Video, point trajectories, and scene geometry require a prediction-to-execution interface. Figure~\ref{fig:teaser} frames this gap between prediction and execution. Geometric fitting can convert predicted motion into a terminal goal, but future prediction training does not optimize recovered goal accuracy. Prediction errors can still produce inaccurate control targets. Figure~\ref{fig:motivation} illustrates inconsistencies in generated futures. Our comparisons show that terminal goal errors remain after predictor training losses converge. Converged prediction objectives do not guarantee accurate goals for control.

\begin{figure*}[t]
\centering
\includegraphics[width=\textwidth]{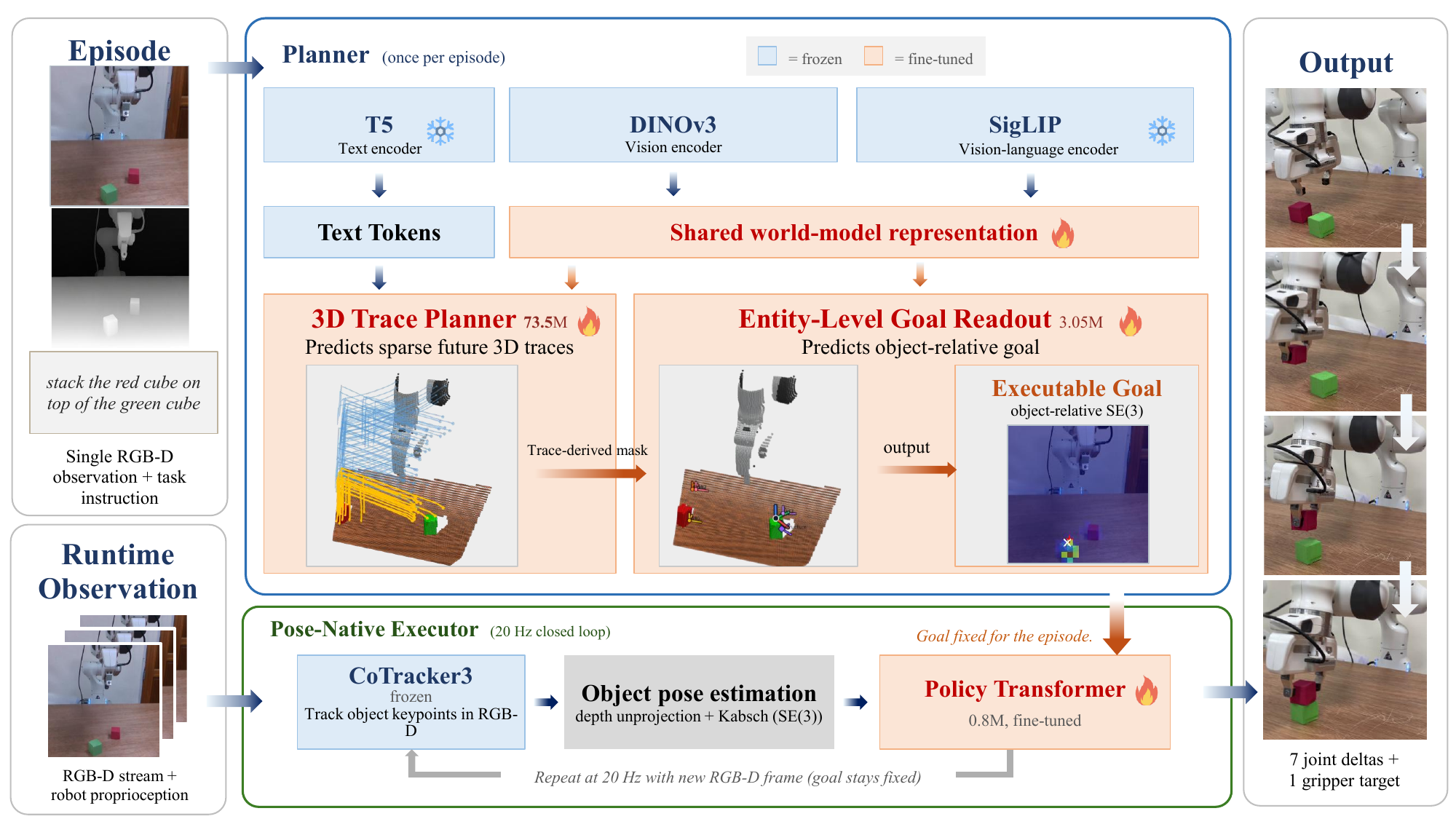}%
\par\vspace{-5pt}
\caption{\textbf{System overview.} Frozen encoders supply a shared world-model representation for trace prediction and Entity-Level Goal Readout. The readout exposes one object-relative goal, anchored at $t=0$ and fixed for the episode. Online tracking and pose estimation update object state to close the loop. The Pose-Native Executor combines pose, goal, proprioception, and previous action to produce an eight-dimensional arm and gripper command at $20$\,Hz.}
\label{fig:pipeline}
\end{figure*}

The design question is not whether a conversion procedure exists, but whether the task variables consumed by control are explicitly learned and exposed. We make terminal goal prediction part of the learning objective through object-centric prediction and observed geometry. The prediction-to-execution interface is learned as part of the model rather than left to incidental post-processing. We present \emph{Entity-Level Goal Readout}, a learned interface between a 3D trace world model and a Pose-Native Executor. Figure~\ref{fig:pipeline} traces this interface from the world-model representation to closed-loop execution. An entity branch predicts terminal object pose, while a map branch grounds translation in observed depth to construct a compact executable goal in $SE(3)$. A shared executor combines this goal with pose feedback to correct execution error without rerunning the world model. This separates episodic planning from continuous control while preserving the task variables required by control.

We evaluate the interface through goal accuracy, manipulation success, and sensitivity to goal error. Across five tasks, our pipeline achieves higher mean success than both rigid readout pipelines under shared executor weights, with the largest gains on PickCube and StackCube. Goal error distributions characterize the targets delivered by the planner, while controlled translation perturbations measure the fixed executor's response to imposed goal error. Real robot trials extend evaluation to distractors and a target unseen during policy training. Our contributions are threefold:

\textbf{A Learned Goal Interface.} Entity-Level Goal Readout makes control-relevant terminal variables explicit learning targets by combining object-centric pose prediction with depth-grounded spatial readout to expose executable goals from the world-model representation.

\textbf{Execution Trained for Imperfect Goals.} A shared Pose-Native Executor is trained under measured planner goal errors using privileged teachers and online pose feedback, without rerunning the planner or receiving privileged guidance during execution.

\textbf{An Empirical Account of Goal Usability.} Goal diagnostics across prediction representations, shared-executor evaluations, controlled perturbations, and real robot trials provide an empirical account of how goal quality translates into closed-loop execution.


\section{Related Work}
\label{sec:related}

\textbf{Generative World Models.} World model planners represent future motion as generated observations, point trajectories, or scene geometry. Dream2Flow~\cite{dream2flow2025} and TesserAct~\cite{tesseract2025} recover motion or object goals from generated observations. ATM~\cite{atm2024} and Track2Act~\cite{track2act2024} predict 2D point trajectories, with Track2Act recovering geometric poses for control. TraceGen~\cite{tracegen2026} and $\mu_0$~\cite{mu0_2026} predict 3D traces, PointWorld~\cite{pointworld2026} predicts action-conditioned 3D motion, and Structured-4D~\cite{structure4d2026} forecasts scene geometry. Across these representations, our focus is how predicted futures become task goals consumed by control.

\textbf{From Predicted Futures to Executable Goals.} Geometric readouts recover control targets through pose estimation or rigid registration, including Kabsch fitting with random sample consensus (RANSAC)~\cite{kabsch1976},~\cite{ransac1981}. Dream2Flow~\cite{dream2flow2025} connects video generation to control through 3D object flow. Such interfaces determine how predicted futures are exposed to control, but future prediction supervision alone does not directly optimize terminal goal accuracy. Our readout makes the terminal goal an explicit learning target and evaluates the resulting goal quality in Table~\ref{tab:goalerr}.

\textbf{Goal-Conditioned Control.} Diffusion Policy~\cite{diffusionpolicy2024}, DP3~\cite{dp3_2024}, and vision language action models map rich observations to actions. Our Pose-Native Executor instead receives one fixed pose goal in $SE(3)$ per episode. Dataset aggregation and privileged teacher methods motivate its distillation strategy~\cite{dagger2011},~\cite{chen2020learningbycheating},~\cite{lee2020legged}. This keeps the prediction-to-execution handoff compact while separating goal quality from closed-loop execution.

\section{Method}
\label{sec:method}


\begin{figure*}[t]
\centering
\includegraphics[width=\textwidth]{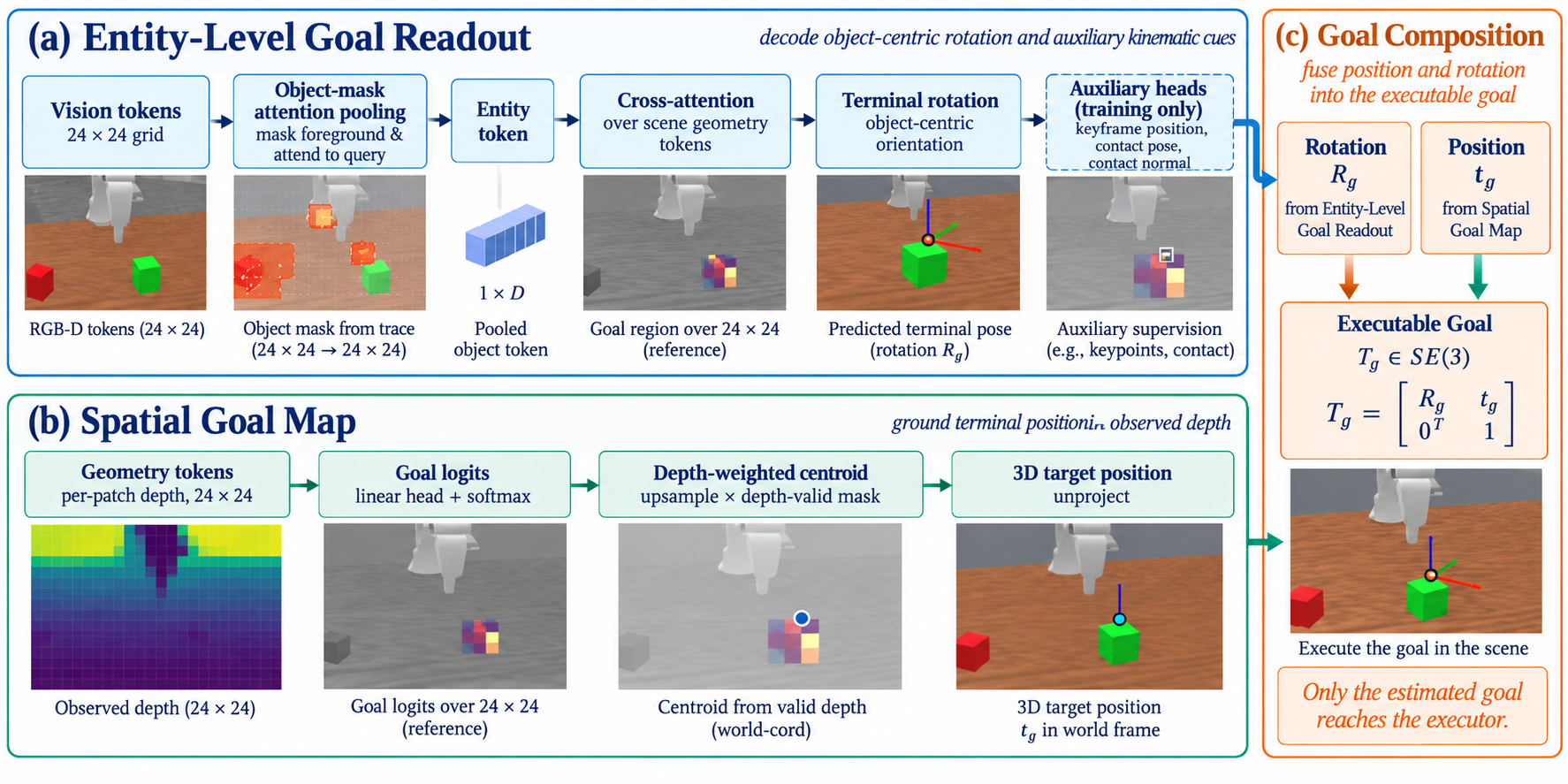}%
\par\vspace{-5pt}
\caption{\textbf{Entity-Level Goal Readout.}
(a) Masked pooling and cross attention produce terminal rotation, with
dashed paths for training-only auxiliary supervision.
(b) The map branch grounds position in valid observed depth.
(c) Goal composition combines rotation and position into the executable
goal. Image panels illustrate intermediate representations.}
\label{fig:readout}
\end{figure*}

\subsection{Overview and Problem Statement}
\label{sec:overview}

The Entity-Level Goal Readout provides a learned prediction-to-execution interface between a 3D trace world model and goal-conditioned control, as shown in Fig.~\ref{fig:pipeline}. The system separates planning once per episode from closed-loop execution. Given a single RGB-D observation and task instruction, the world model predicts future scene motion. From the shared world model representation, the interface explicitly predicts one executable goal $\Delta\hat T_g \in SE(3)$ relative to the initial object frame.

Execution runs independently at $20$\,Hz. An online perception module estimates the current object pose $\hat T_o(t)$ in the world frame. A compact Pose-Native Executor maps the current pose, fixed executable goal, robot proprioception, and previous action to commands for seven arm joint increments and a gripper position target. The planner communicates a compact seven-parameter representation of the $SE(3)$ goal rather than the full predicted future. The planner and rigid baseline are described in Sec.~\ref{sec:planner}, the readout in Sec.~\ref{sec:readout}, and the executor and its training in Sec.~\ref{sec:executor}. Task success is evaluated separately from goal accuracy, as described in Sec.~\ref{sec:exp_protocol}.

\subsection{Trace Planner and Rigid Readout}
\label{sec:planner}

\textbf{Backbone.} We build on TraceGen~\cite{tracegen2026}, which predicts sparse 3D point trajectories rather than rendered future frames. DINOv3~\cite{dinov3_2025} and SigLIP~\cite{siglip2023} features are arranged on a $24\times24$ spatial grid and augmented with depth, while a frozen T5~\cite{t5_2020} encoder embeds the task instruction. A flow matching decoder based on CogVideoX~\cite{cogvideox2024} predicts trajectories for $400$ query points over $32$ future steps.

The visual and language encoders remain frozen while planner layers adapted to each task are fine tuned on manipulation demonstrations. Evaluation uses the checkpoints and readout routes specified in Sec.~\ref{sec:exp_interface}.

\textbf{Rigid Readout.} The conventional \emph{rigid readout} recovers a terminal transform from predicted object tracks using weighted Kabsch~\cite{kabsch1976} inside RANSAC~\cite{ransac1981}. This provides a strong geometric baseline, but the underlying predictor is not directly trained for terminal goal accuracy, and the task goal must be recoverable as a single rigid transform. Here $K$ denotes the number of stochastic trace predictions. For $K=4$, traces are averaged before rigid fitting rather than selecting or averaging solved poses.

\subsection{Entity-Level Goal Readout}
\label{sec:readout}

Figure~\ref{fig:readout} details the two branches that explicitly predict task relevant quantities from the shared planner representation.

\textbf{Entity Branch.} An object mask derived from motion selects the relevant support on the $24\times24$ vision token grid. A learned query performs masked attention pooling to produce a single entity token. Cross attention to the $576$ scene tokens enriched with geometry yields an object representation from which we predict the terminal object pose, whose rotation is used for executable goal composition. Keyframe poses, contact point, and contact normal are predicted only as auxiliary supervision and are not passed to the executor.

\textbf{Map Branch.} Terminal position is grounded directly in observed scene geometry. From the same spatial geometry tokens, a linear head predicts logits over a $24\times24$ goal map. After softmax and upsampling, the distribution is masked by valid observed depth and reduced to an image centroid weighted by depth. Unprojection through the RGB-D camera model yields a camera frame point. The calibrated camera to world transform $T_{WC}$ maps it to the world position $\hat{\mathbf t}_g$. This readout also supports targets defined by regions that do not correspond to a unique rigid transform.

\textbf{Goal Composition.} Eq.~\eqref{eq:goal} combines the map position with the entity rotation, both expressed in the world frame, to form the terminal object pose.
\begin{equation}
\hat T_g =
\begin{bmatrix}
\hat R_g & \hat{\mathbf t}_g \\
\mathbf 0^\top & 1
\end{bmatrix}
\in SE(3),
\label{eq:goal}
\end{equation}
The interface expresses this pose relative to the initial object frame as $\Delta\hat T_g=\hat T_o(0)^{-1}\hat T_g$. Here $\hat T_o(0)$ maps initial object coordinates to world coordinates.

\textbf{Training Objective.} The readout is trained jointly with the trace objective in Eq.~\eqref{eq:loss}.
\begin{equation}
\mathcal{L}
=
\mathcal{L}_{\mathrm{flow}}
+\lambda_{\mathrm{ent}}\mathcal{L}_{\mathrm{ent}}
+\lambda_{\mathrm{map}}\mathcal{L}_{\mathrm{map}},
\label{eq:loss}
\end{equation}
where $\mathcal{L}_{\mathrm{ent}}$ supervises terminal pose and auxiliary interaction quantities and $\mathcal{L}_{\mathrm{map}}$ supervises the spatial goal distribution. These terms directly supervise executable goal prediction rather than relying on goals recovered from future traces after prediction.


\begin{table*}[!t]
\centering
\caption{\textbf{Terminal goal position error} in mm, reported as mean
$\pm$ standard deviation over valid readouts. Source-specific standard
deviation conventions are retained and documented in the supplementary
appendix. Lower is better.
Evaluation banks and validity conventions are specified
in Sec.~\ref{sec:exp_protocol}.}
\label{tab:goalerr}
\small
\setlength{\tabcolsep}{12pt}
\renewcommand{\arraystretch}{1.12}
\begin{tabular}{@{}llccc@{}}
\toprule
Prediction & Method & PickCube & PegInsertionSide & StackCube \\
\midrule
RGB video & Cosmos~3 Edge
& $141.3\!\pm\!89.3$ & $348.2\!\pm\!214.9$ & $87.6\!\pm\!61.6$ \\
RGB-DN video & TesserAct, adapted
& $456.9\!\pm\!158.3$ & $76.9\!\pm\!38.2$ & $224.8\!\pm\!89.0$ \\
2D tracks & Track2Act, adapted
& $167.5\!\pm\!57.0$ & $256.2\!\pm\!33.0$ & $143.0\!\pm\!89.0$ \\
4D geometry & Structured-4D, released
& $192.2\!\pm\!82.9$ & $373.5\!\pm\!115.3$ & $31.0\!\pm\!19.3$ \\
4D geometry & Structured-4D, adapted
& $166.4\!\pm\!61.3$ & $361.4\!\pm\!83.3$ & $36.3\!\pm\!20.3$ \\
\midrule
3D tracks & TraceGen + Rigid Readout ($K=1$)
& $41.1\!\pm\!21.7$ & $42.6\!\pm\!31.6$ & $28.9\!\pm\!38.7$ \\
3D tracks & TraceGen + Rigid Readout ($K=4$)
& $37.2\!\pm\!19.3$ & $40.8\!\pm\!33.0$ & $25.4\!\pm\!38.8$ \\
3D tracks + goal & \textbf{Entity-Level Goal Readout}
& $\mathbf{31.5\!\pm\!20.1}$ & $\mathbf{29.4\!\pm\!16.4}$
& $\mathbf{12.8\!\pm\!34.0}$ \\
\bottomrule
\end{tabular}
\end{table*}

\subsection{Executor and Distillation}
\label{sec:executor}

The readout runs once at episode start. The executor reconstructs the fixed world goal as $\hat T_g=\hat T_o(0)\Delta\hat T_g$ and compares it with the online world pose $\hat T_o(t)$. Both use the same world frame. The goal remains fixed while pose feedback updates the current state, correcting execution without rerunning the world model.

\textbf{Online Perception.} An initial object mask supplies tracking points in RGB-D observations. Frozen CoTracker3~\cite{cotracker3_2025} maintains their image correspondences. Depth unprojection followed by Kabsch recovers $\hat T_o(t)$, which is updated while the executable goal remains fixed.

\textbf{Policy.} The compact 0.8M-parameter transformer receives the current object pose, executable goal, joint positions and velocities, tool center point pose, and previous action. Its eight dimensional output comprises seven arm joint increments and a gripper position target at $20$\,Hz. The same student policy is used across all five tasks. Current pose feedback supports correction during execution, while the fixed goal decouples control updates from planner inference.

\textbf{Distillation.} We use dataset aggregation (DAgger)~\cite{dagger2011}. The student controls the environment, and the privileged task policy supplies action labels at the visited states. The teacher additionally receives a privileged guidance variable $\psi$, which is not an input to the student during training or deployment. Goal perturbations are sampled from the measured planner error distribution during student training, exposing the controller to imperfect goals rather than only designated correct goals. Privileged guidance therefore enters through teacher supervision, not through the student's observation at execution.

\section{Experiments}
\label{sec:experiments}

We evaluate the prediction-to-execution interface through terminal goal quality across prediction representations, closed-loop success under a shared executor, and controller sensitivity to imposed goal error. Real robot trials extend the evaluation beyond simulation.

\textbf{Evaluation Protocol.}
\label{sec:exp_protocol}
We use five ManiSkill3~\cite{maniskill3_2025} tasks. Position comparisons focus on PickCube, PegInsertionSide, and StackCube. LiftPegUpright admits multiple upright positions and PushCube accepts a broad target region, so both remain in success evaluations. Archived rigid and Oracle results pool three evaluation seeds with $256$ scenes each, yielding $768$ episodes per task and goal source. Pose diagnostics use $256$ matched scenes. Perturbations use two evaluation seeds and $512$ episodes per point. These protocols are separate from policy training. Simulation training and timing use a single RTX~5090.

Table~\ref{tab:goalerr} measures 3D Euclidean distance between recovered terminal goal positions and reference goal positions, not image prediction or scene reconstruction error. It uses the Structured-4D evaluation bank, the diagnostic bank for the other external planners including adapted Track2Act, and stored goals for the final three rows. Statistics include finite estimates marked unsolved. The table and discussion use all valid readouts for each model rather than a common valid subset across checkpoints.

\textbf{Baselines.}
Table~\ref{tab:goalerr} groups planners by RGB or RGB-DN video, 2D or 3D point trajectories, and 3D scene evolution. The final three rows use the same 3D point trajectory representation and compare rigid recovery with learned goal readout, focusing the comparison on how predicted futures are exposed to control. Table~\ref{tab:main} separates native executors from the shared executor comparison. Oracle Goal rollouts use designated correct goals.

\subsection{Goal Quality Across Planners}
\label{sec:exp_goal_error}

\textbf{Fine Tuning.}
All three baseline adaptations run on a single RTX~5090 across five tasks. TesserAct generates RGB-DN video and uses low rank adaptation (LoRA) on $4{,}962$ clips. Structured-4D predicts evolving 3D scene geometry with a two-stage generator adapted using LoRA on $9{,}120$ frame pairs. Track2Act predicts 2D point trajectories and uses full model fine tuning on $4{,}493$ clips. Training losses converge, but recovered terminal goals retain substantial errors. Training data and optimization budgets differ across methods.

Structured-4D reconstructs geometry at roughly centimeter scale, yet substantial PickCube and PegInsertionSide goal errors remain. Adapted TesserAct has lower position error on PegInsertionSide than on PickCube or StackCube, but substantial errors remain. Convergence under the prediction objective does not ensure accurate terminal goals. Within the shared 3D trace representation, Entity-Level Goal Readout gives the lowest mean position error on all three reported tasks. Its reported errors are $15.3\%$, $27.9\%$, and $49.6\%$ lower than rigid $K=4$ on PickCube, PegInsertionSide, and StackCube, respectively (Table~\ref{tab:goalerr}).

\subsection{Closed-Loop Execution}
\label{sec:exp_interface}


\begin{table*}[!t]
\centering
\caption{
\textbf{Closed-loop task success.}
External systems use native executors. The shared-executor block uses the
same policy weights and paired scenes, but different goal pipelines and
planner checkpoints. Native-executor results are separate system-level
evaluations. Rates are percentages. Mean is the unweighted average of all
five displayed task rates, rounded to two decimal places. It is omitted
for incomplete rows. Bold and underline
mark the best and second-best results in each column across both blocks,
ranked using the reported values. Dashes denote unreported results.
Oracle Goal references are shown separately in Fig.~\ref{fig:causal}(a).
}
\label{tab:main}
\small
\setlength{\tabcolsep}{4pt}
\setlength{\dashlinedash}{1.5pt}
\setlength{\dashlinegap}{2pt}
\renewcommand{\arraystretch}{1.15}

\begin{tabular}{@{}lccccc:c@{}}
\toprule
& \multicolumn{6}{c}{Success (\%)} \\
Method & PickCube & LiftPegUpright & PegInsertionSide & StackCube & PushCube & Mean \\
\midrule

\multicolumn{1}{l}{\textit{Native executor}} & & & & & & \\

DP3~\cite{dp3_2024}
& 36.00
& \underline{84.00}
& 12.00
& 20.00
& 93.00
& 49.00 \\

RoboFlow4D~\cite{roboflow4d2026}
& 14.00
& --
& --
& 12.00
& 45.00
& -- \\

Structured-4D, released
& 0.00
& 0.00
& 16.00
& 84.00
& 0.00
& 20.00 \\

Structured-4D, adapted
& 33.00
& 16.00
& 20.00
& \textbf{89.00}
& 92.00
& 50.00 \\

Cosmos~3 Edge Policy, adapted
& 3.00
& 0.00
& 0.00
& 13.33
& 23.33
& 7.93 \\

\midrule

\multicolumn{1}{l}{\textit{Shared executor}} & & & & & & \\

Rigid Readout ($K=1$)
& 29.80
& 70.18
& 20.05
& 52.47
& \textbf{99.74}
& 54.45 \\

Rigid Readout ($K=4$)
& \underline{46.48}
& 76.95
& \underline{21.88}
& 54.82
& \underline{99.22}
& \underline{59.87} \\

\textbf{Entity-Level Goal Readout}
& \textbf{81.50}
& \textbf{98.35}
& \textbf{32.84}
& \underline{86.54}
& 99.20
& \textbf{79.69} \\

\bottomrule
\end{tabular}
\end{table*}

\begin{figure*}[!t]
\centering
\resizebox{\textwidth}{!}{\input{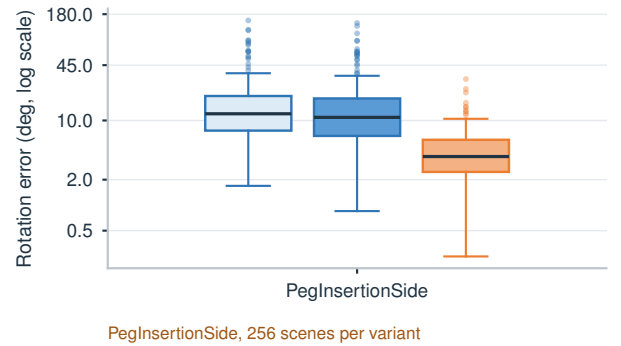}}%
\par\vspace{-5pt}
\caption{\textbf{Task success and goal error distributions.}
(a) Light blue, blue, and orange denote rigid $K=1$, rigid $K=4$, and the
final pipeline. Gray segments show empirical Oracle Goal references.
(b,c) Seed 999 box plots show the median and interquartile range (IQR),
$1.5$ IQR whiskers, and all
outliers on logarithmic axes. Groups have $256$ finite records except
StackCube rigid positions ($252/256$). (c) Raw quaternion errors are shown
for PegInsertionSide. Comparison conditions are in
Sec.~\ref{sec:exp_interface}.}
\label{fig:causal}
\end{figure*}

\textbf{Comparison Scope.}
Shared-executor comparisons use different checkpoints and samples, one policy-training seed, and noise matched to final-pipeline errors. Rigid readout uses the earlier trace checkpoint for four tasks. Final routing uses the goal-map checkpoint for PickCube and StackCube and position-head banks elsewhere. This supports pipeline comparison, not isolated readout attribution. Native executors provide separate system-level context.

Under shared executor weights, the final pipeline achieves $79.69\%$ mean success, compared with $54.45\%$ for $K=1$ and $59.87\%$ for $K=4$ (Table~\ref{tab:main}). Its mean success is $19.82$ points higher than $K=4$. The final pipeline exceeds both rigid variants on four tasks, with the largest gains over $K=4$ on PickCube and StackCube. PushCube remains near saturation at $99.20\%$, slightly below both rigid variants. Additional trace averaging improves rigid readout on most tasks but does not close the mean success gap.

Adapted Structured-4D achieves higher StackCube success ($89.00\%$ versus $86.54\%$, Table~\ref{tab:main}), but uses weights trained with $1{,}000$ demonstrations per task and its native executor. Its 1.1B planner takes 4.45\,s per plan versus 1.27\,s for our 677.6M planner (Table~\ref{tab:cost}), while our executor is shared across five tasks.

\textbf{Oracle Reference.}
\label{sec:exp_oracle}
Gray segments in Fig.~\ref{fig:causal}(a) indicate execution difficulty even with designated correct goals. The updated LiftPegUpright result exceeds its archived Oracle reference, so these measurements remain empirical reference points rather than theoretical upper bounds.

\subsection{Readout Diagnostics}
\label{sec:exp_pose_quality}

\begin{figure}[!t]
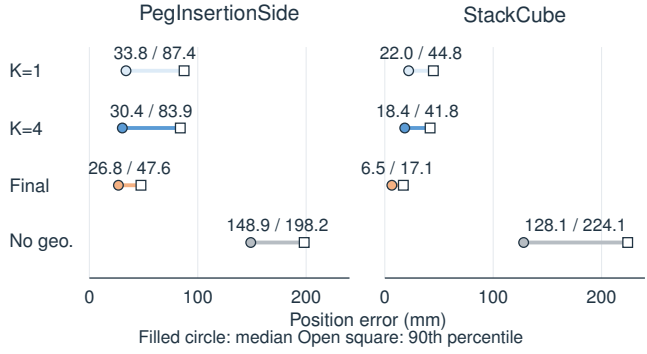

\centering
\resizebox{\columnwidth}{!}{%
\begin{pgfpicture}
\pgfpathrectangle{\pgfpointorigin}{\pgfqpoint{3.5in}{1.9200in}}
\pgfusepath{use as bounding box}
\definecolor{ink}{HTML}{203342}
\definecolor{lightblue}{HTML}{DDEBF7}
\definecolor{blue}{HTML}{5B9BD5}
\definecolor{orange}{HTML}{F4B183}
\definecolor{gray}{HTML}{B7BDC3}
\definecolor{grid}{HTML}{E5E9ED}
\pgftext[x=1.15000in,y=1.77844in,]{\color{ink}\sffamily\fontsize{8.2}{9.84}\selectfont PegInsertionSide}
\pgfsetstrokecolor{grid}
\pgfsetlinewidth{0.4pt}
\pgfpathmoveto{\pgfqpoint{0.45000in}{0.36000in}}
\pgfpathlineto{\pgfqpoint{0.45000in}{1.60313in}}
\pgfusepath{stroke}
\pgftext[x=0.45000in,y=.26000in,]{\color{ink}\sffamily\fontsize{7}{8.4}\selectfont 0}
\pgfsetstrokecolor{grid}
\pgfsetlinewidth{0.4pt}
\pgfpathmoveto{\pgfqpoint{1.03334in}{0.36000in}}
\pgfpathlineto{\pgfqpoint{1.03334in}{1.60313in}}
\pgfusepath{stroke}
\pgftext[x=1.03334in,y=.26000in,]{\color{ink}\sffamily\fontsize{7}{8.4}\selectfont 100}
\pgfsetstrokecolor{grid}
\pgfsetlinewidth{0.4pt}
\pgfpathmoveto{\pgfqpoint{1.61666in}{0.36000in}}
\pgfpathlineto{\pgfqpoint{1.61666in}{1.60313in}}
\pgfusepath{stroke}
\pgftext[x=1.61666in,y=.26000in,]{\color{ink}\sffamily\fontsize{7}{8.4}\selectfont 200}
\pgfsetstrokecolor{ink}
\pgfsetlinewidth{0.5pt}
\pgfpathmoveto{\pgfqpoint{0.45000in}{0.36000in}}
\pgfpathlineto{\pgfqpoint{1.85000in}{0.36000in}}
\pgfusepath{stroke}
\pgftext[x=0.02000in,y=1.47562in,left]{\color{ink}\sffamily\fontsize{7.2}{8.64}\selectfont K=1}
\pgfsetstrokecolor{lightblue}
\pgfsetlinewidth{1.5pt}
\pgfpathmoveto{\pgfqpoint{0.64716in}{1.47562in}}
\pgfpathlineto{\pgfqpoint{0.95984in}{1.47562in}}
\pgfusepath{stroke}
\pgfsetstrokecolor{ink}
\pgfsetfillcolor{lightblue}
\pgfsetlinewidth{0.5pt}
\pgfpathcircle{\pgfqpoint{0.64716in}{1.47562in}}{.026in}
\pgfusepath{stroke,fill}
\pgfsetfillcolor{white}
\pgfpathrectangle{\pgfqpoint{0.93384in}{1.44963in}}{\pgfqpoint{.052in}{.052in}}
\pgfusepath{stroke,fill}
\pgftext[x=0.80350in,y=1.57125in,]{\color{ink}\sffamily\fontsize{6.8}{8.16}\selectfont 33.8 / 87.4}
\pgftext[x=0.02000in,y=1.16750in,left]{\color{ink}\sffamily\fontsize{7.2}{8.64}\selectfont K=4}
\pgfsetstrokecolor{blue}
\pgfsetlinewidth{1.5pt}
\pgfpathmoveto{\pgfqpoint{0.62734in}{1.16750in}}
\pgfpathlineto{\pgfqpoint{0.93942in}{1.16750in}}
\pgfusepath{stroke}
\pgfsetstrokecolor{ink}
\pgfsetfillcolor{blue}
\pgfsetlinewidth{0.5pt}
\pgfpathcircle{\pgfqpoint{0.62734in}{1.16750in}}{.026in}
\pgfusepath{stroke,fill}
\pgfsetfillcolor{white}
\pgfpathrectangle{\pgfqpoint{0.91342in}{1.14150in}}{\pgfqpoint{.052in}{.052in}}
\pgfusepath{stroke,fill}
\pgftext[x=0.78338in,y=1.26313in,]{\color{ink}\sffamily\fontsize{6.8}{8.16}\selectfont 30.4 / 83.9}
\pgftext[x=0.02000in,y=0.85938in,left]{\color{ink}\sffamily\fontsize{7.2}{8.64}\selectfont Final}
\pgfsetstrokecolor{orange}
\pgfsetlinewidth{1.5pt}
\pgfpathmoveto{\pgfqpoint{0.60634in}{0.85938in}}
\pgfpathlineto{\pgfqpoint{0.72766in}{0.85938in}}
\pgfusepath{stroke}
\pgfsetstrokecolor{ink}
\pgfsetfillcolor{orange}
\pgfsetlinewidth{0.5pt}
\pgfpathcircle{\pgfqpoint{0.60634in}{0.85938in}}{.026in}
\pgfusepath{stroke,fill}
\pgfsetfillcolor{white}
\pgfpathrectangle{\pgfqpoint{0.70166in}{0.83337in}}{\pgfqpoint{.052in}{.052in}}
\pgfusepath{stroke,fill}
\pgftext[x=0.66700in,y=0.95500in,]{\color{ink}\sffamily\fontsize{6.8}{8.16}\selectfont 26.8 / 47.6}
\pgftext[x=0.02000in,y=0.55125in,left]{\color{ink}\sffamily\fontsize{7.2}{8.64}\selectfont No geo.}
\pgfsetstrokecolor{gray}
\pgfsetlinewidth{1.5pt}
\pgfpathmoveto{\pgfqpoint{1.31858in}{0.55125in}}
\pgfpathlineto{\pgfqpoint{1.60616in}{0.55125in}}
\pgfusepath{stroke}
\pgfsetstrokecolor{ink}
\pgfsetfillcolor{gray}
\pgfsetlinewidth{0.5pt}
\pgfpathcircle{\pgfqpoint{1.31858in}{0.55125in}}{.026in}
\pgfusepath{stroke,fill}
\pgfsetfillcolor{white}
\pgfpathrectangle{\pgfqpoint{1.58016in}{0.52525in}}{\pgfqpoint{.052in}{.052in}}
\pgfusepath{stroke,fill}
\pgftext[x=1.46238in,y=0.64687in,]{\color{ink}\sffamily\fontsize{6.8}{8.16}\selectfont 148.9 / 198.2}
\pgftext[x=2.74000in,y=1.77844in,]{\color{ink}\sffamily\fontsize{8.2}{9.84}\selectfont StackCube}
\pgfsetstrokecolor{grid}
\pgfsetlinewidth{0.4pt}
\pgfpathmoveto{\pgfqpoint{2.04000in}{0.36000in}}
\pgfpathlineto{\pgfqpoint{2.04000in}{1.60313in}}
\pgfusepath{stroke}
\pgftext[x=2.04000in,y=.26000in,]{\color{ink}\sffamily\fontsize{7}{8.4}\selectfont 0}
\pgfsetstrokecolor{grid}
\pgfsetlinewidth{0.4pt}
\pgfpathmoveto{\pgfqpoint{2.62334in}{0.36000in}}
\pgfpathlineto{\pgfqpoint{2.62334in}{1.60313in}}
\pgfusepath{stroke}
\pgftext[x=2.62334in,y=.26000in,]{\color{ink}\sffamily\fontsize{7}{8.4}\selectfont 100}
\pgfsetstrokecolor{grid}
\pgfsetlinewidth{0.4pt}
\pgfpathmoveto{\pgfqpoint{3.20666in}{0.36000in}}
\pgfpathlineto{\pgfqpoint{3.20666in}{1.60313in}}
\pgfusepath{stroke}
\pgftext[x=3.20666in,y=.26000in,]{\color{ink}\sffamily\fontsize{7}{8.4}\selectfont 200}
\pgfsetstrokecolor{ink}
\pgfsetlinewidth{0.5pt}
\pgfpathmoveto{\pgfqpoint{2.04000in}{0.36000in}}
\pgfpathlineto{\pgfqpoint{3.44000in}{0.36000in}}
\pgfusepath{stroke}
\pgfsetstrokecolor{lightblue}
\pgfsetlinewidth{1.5pt}
\pgfpathmoveto{\pgfqpoint{2.16834in}{1.47562in}}
\pgfpathlineto{\pgfqpoint{2.30134in}{1.47562in}}
\pgfusepath{stroke}
\pgfsetstrokecolor{ink}
\pgfsetfillcolor{lightblue}
\pgfsetlinewidth{0.5pt}
\pgfpathcircle{\pgfqpoint{2.16834in}{1.47562in}}{.026in}
\pgfusepath{stroke,fill}
\pgfsetfillcolor{white}
\pgfpathrectangle{\pgfqpoint{2.27534in}{1.44963in}}{\pgfqpoint{.052in}{.052in}}
\pgfusepath{stroke,fill}
\pgftext[x=2.23484in,y=1.57125in,]{\color{ink}\sffamily\fontsize{6.8}{8.16}\selectfont 22.0 / 44.8}
\pgfsetstrokecolor{blue}
\pgfsetlinewidth{1.5pt}
\pgfpathmoveto{\pgfqpoint{2.14734in}{1.16750in}}
\pgfpathlineto{\pgfqpoint{2.28384in}{1.16750in}}
\pgfusepath{stroke}
\pgfsetstrokecolor{ink}
\pgfsetfillcolor{blue}
\pgfsetlinewidth{0.5pt}
\pgfpathcircle{\pgfqpoint{2.14734in}{1.16750in}}{.026in}
\pgfusepath{stroke,fill}
\pgfsetfillcolor{white}
\pgfpathrectangle{\pgfqpoint{2.25784in}{1.14150in}}{\pgfqpoint{.052in}{.052in}}
\pgfusepath{stroke,fill}
\pgftext[x=2.21558in,y=1.26313in,]{\color{ink}\sffamily\fontsize{6.8}{8.16}\selectfont 18.4 / 41.8}
\pgfsetstrokecolor{orange}
\pgfsetlinewidth{1.5pt}
\pgfpathmoveto{\pgfqpoint{2.07792in}{0.85938in}}
\pgfpathlineto{\pgfqpoint{2.13974in}{0.85938in}}
\pgfusepath{stroke}
\pgfsetstrokecolor{ink}
\pgfsetfillcolor{orange}
\pgfsetlinewidth{0.5pt}
\pgfpathcircle{\pgfqpoint{2.07792in}{0.85938in}}{.026in}
\pgfusepath{stroke,fill}
\pgfsetfillcolor{white}
\pgfpathrectangle{\pgfqpoint{2.11375in}{0.83337in}}{\pgfqpoint{.052in}{.052in}}
\pgfusepath{stroke,fill}
\pgftext[x=2.10884in,y=0.95500in,]{\color{ink}\sffamily\fontsize{6.8}{8.16}\selectfont 6.5 / 17.1}
\pgfsetstrokecolor{gray}
\pgfsetlinewidth{1.5pt}
\pgfpathmoveto{\pgfqpoint{2.78724in}{0.55125in}}
\pgfpathlineto{\pgfqpoint{3.34724in}{0.55125in}}
\pgfusepath{stroke}
\pgfsetstrokecolor{ink}
\pgfsetfillcolor{gray}
\pgfsetlinewidth{0.5pt}
\pgfpathcircle{\pgfqpoint{2.78724in}{0.55125in}}{.026in}
\pgfusepath{stroke,fill}
\pgfsetfillcolor{white}
\pgfpathrectangle{\pgfqpoint{3.32125in}{0.52525in}}{\pgfqpoint{.052in}{.052in}}
\pgfusepath{stroke,fill}
\pgftext[x=3.06724in,y=0.64687in,]{\color{ink}\sffamily\fontsize{6.8}{8.16}\selectfont 128.1 / 224.1}
\pgftext[x=1.95000in,y=0.13000in,]{\color{ink}\sffamily\fontsize{7.2}{8.64}\selectfont Position error (mm)}
\pgftext[x=1.75000in,y=0.03500in,]{\color{ink}\sffamily\fontsize{6.8}{8.16}\selectfont Filled circle: median     Open square: 90th percentile}
\end{pgfpicture}}
\caption{\textbf{Position Quality and Geometry Probe.}
Median / 90th-percentile errors in mm. Lower is better.
Final denotes the routed pipeline. No geo. removes geometry from the
entity-position predictor at test time. It is not an ablation of Final.
Segments connect
quantiles, not confidence bounds. All groups use $256$ scenes, with
$252$ finite StackCube rigid positions and $256$ otherwise.}
\label{fig:geometry_probe}
\end{figure}

The final pipeline reduces both typical and tail position error on StackCube in Fig.~\ref{fig:causal}(b) and Fig.~\ref{fig:geometry_probe}, while PegInsertionSide shows a clearer tail reduction. Quantiles exclude nonfinite estimates, but execution denominators retain all scenes. The $25$\,mm StackCube proxy improves even when finite unsolved goals count as failures, whereas PegInsertionSide's $20$\,mm proxy changes little, showing that a single threshold does not fully characterize goal quality.

Raw quaternion errors in Fig.~\ref{fig:causal}(c) are lower for the final PegInsertionSide bank, which uses the entity-branch pose prediction. Removing geometry from the entity-position head increases error in Fig.~\ref{fig:geometry_probe}, indicating reliance on scene geometry in this probe rather than a full training-loss ablation. Quaternion errors are not symmetry adjusted and do not establish rotation-mediated execution gains. Final and $K=4$ LiftPegUpright rotations are identical, while its upright predicate and StackCube symmetry require task-aware metrics, so no learned-rotation gain is claimed for either task.

\subsection{Position Sensitivity}
\label{sec:exp_sensitivity}


\begin{figure}[!t]
\centering
\resizebox{\columnwidth}{!}{\input{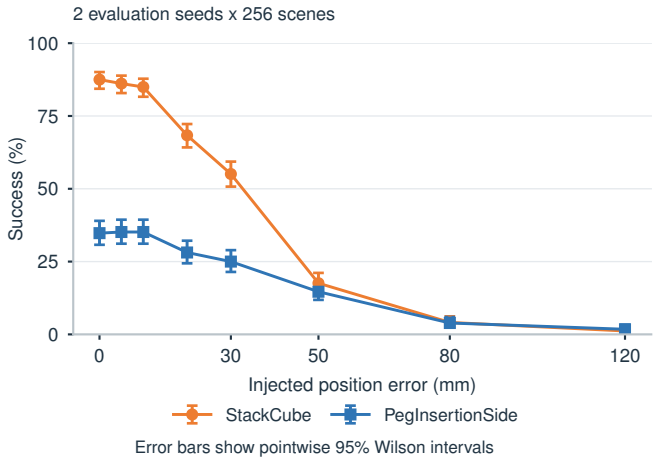}}%
\par\vspace{-5pt}
\caption{\textbf{Execution sensitivity to goal position.}
The Oracle Goal is translated with the planner bypassed. Each point pools
$256$ scenes at seeds 999 and 997 ($512$ episodes), with scenes and
directions paired across magnitudes. Markers show measured success,
error bars show pointwise $95\%$ Wilson intervals, and lines guide the eye.
PegInsertionSide uses $10$\,mm clearance.}
\label{fig:sensitivity}
\end{figure}

We bypass the planner and translate the Oracle Goal along directions sampled isotropically in the world frame, including vertical offsets. Offsets stay fixed per episode, with scenes and directions paired across magnitudes. Only each environment's initial episode is scored. The two tasks span $8{,}192$ rollouts over eight settings. PegInsertionSide uses widened $10$\,mm clearance. Error bars are pointwise $95\%$ Wilson intervals, not intervals for paired effects or policy training variability.

Figure~\ref{fig:sensitivity} shows success declining with larger offsets on both tasks, with a sharper StackCube drop from $30$ to $50$\,mm. Each curve retains its measured zero-offset baseline rather than the pipeline Oracle reference. Goal distributions measure errors delivered by a pipeline, while this intervention measures the fixed executor's response to imposed errors. These analyses separate planner-delivered goal error from controller sensitivity, while task success reflects their combined effect. The intervention directly tests whether the accuracy of the task variable exposed to control matters during execution.

\subsection{Computational Cost and Real Robot Deployment}
\label{sec:exp_cost}


\begin{table}[!t]
\centering
\caption{\textbf{Planning cost} on an RTX~5090 after model warmup using the
shared timing audit. TesserAct reports mean $\pm$ standard deviation over
18 clips.}
\label{tab:cost}
\scriptsize
\setlength{\tabcolsep}{0pt}
\begin{tabular*}{\columnwidth}{@{\extracolsep{\fill}}lrrrl@{}}
\toprule
Planner & Parameters & Memory & Time & Output \\
& & (GB) & (s) & \\
\midrule
Cosmos~3 Edge & 9.13B & 19.33 & 35.63 & 81 RGB frames \\
TesserAct & 5.7B & 16.22 & $292.0\!\pm\!4.9$ & 49 RGB-DN frames \\
Track2Act & 469M & 1.91 & 3.80 & 400 2D tracks \\
Structured-4D & 1.1B & 5.71 & 4.45 & 4D point cloud \\
\textbf{Ours} & \textbf{677.6M} & \textbf{5.60} & \textbf{1.27}
& \textbf{3D traces + goal} \\
\bottomrule
\end{tabular*}
\end{table}


\begin{table}[!t]
\centering
\caption{\textbf{Real robot success} on a Franka arm with an RTX~3090.}
\label{tab:real}
\scriptsize
\setlength{\tabcolsep}{0pt}
\begin{tabular*}{\columnwidth}{@{\extracolsep{\fill}}lccc@{}}
\toprule
Task & Setting & Successes / Trials & Rate (\%) \\
\midrule
StackCube & nominal & $22/30$ & 73.33 \\
PickPlate & unseen target & $9/12$ & 75.00 \\
StackCube + distractors & intervention & $8/12$ & 66.67 \\
\bottomrule
\end{tabular*}
\end{table}


\begin{figure*}[!t]
\centering
\resizebox{\textwidth}{!}{\input{figures/generated/3_RealRobot_Revised.tex}}%
\par\vspace{-5pt}
\caption{\textbf{Real robot deployment.}
Rollouts show (a) nominal StackCube, (b) StackCube with distractors, and
(c) PickPlate, where the cube is placed in the orange plate. PickPlate is
unseen during policy training. Frames progress from left to right.}
\label{fig:real_robot}
\end{figure*}

Table~\ref{tab:cost} reports the shared timing audit separately from training-evaluation timing. The readout adds less than 0.5\% to planner parameters. The executor's approximately 1.4\,ms inference fits inside the 20\,Hz loop, while planning runs only once per episode.

\textbf{Real Robot Deployment.}
\label{sec:exp_real}
Franka trials with an RTX~3090 cover nominal StackCube, distractors, and PickPlate (Table~\ref{tab:real}, Fig.~\ref{fig:real_robot}). Nominal StackCube reaches $73.33\%$ success, StackCube with distractors reaches $66.67\%$, and PickPlate reaches $75.00\%$. These trials extend evaluation beyond nominal stacking. PickPlate is unseen during policy training but has a larger valid target region, providing evidence of target transfer rather than greater manipulation difficulty.

\section{Limitations}
\label{sec:limits}

A single terminal pose does not explicitly encode trajectory constraints, contact sequences, or articulated and deformable-object states. Planning occurs once per episode, so pose feedback can correct execution but cannot revise an invalidated goal. The system also depends on reliable tracking, depth, and pose estimates, while new goal semantics may require additional task-relevant supervision. Our results support task-aligned readout rather than the necessity of future prediction, since direct goal prediction may suffice for visually observable objectives.

\section{Conclusion}
\label{sec:conclusion}

We presented Entity-Level Goal Readout, a learned prediction-to-execution interface that makes terminal task variables explicit learning targets through object-centric pose prediction and depth-grounded translation. The reported pipelines improve terminal goal quality and mean closed-loop success, while controlled perturbations show that positional goal error matters to a fixed executor. Real robot trials extend the interface to physical manipulation. The central finding is that rich future predictions do not by themselves expose the task variables required by control. These variables must be explicitly learned and delivered through a reliable prediction-to-execution interface. Our system realizes this principle with terminal pose goals and online feedback. Richer goal representations, goal revision during execution, and uncertainty-aware perception remain directions for broader deployment.

\section*{Acknowledgment}
Claude (Anthropic) and ChatGPT (OpenAI) assisted with manuscript text and figure layouts. The authors produced or verified scientific claims, experimental design, data collection, analysis, and results, assuming full responsibility.

\bibliographystyle{IEEEtran}
\bibliography{refs}

\end{document}